\documentclass[preprint,11pt,nonatbib]{elsarticle}

\usepackage[a4paper,margin=2.5cm]{geometry}
\usepackage{amsmath,amssymb,amsthm}
\usepackage{booktabs}
\usepackage{array}
\usepackage{multirow}
\usepackage{graphicx}
\usepackage{url} 
\usepackage[ruled,vlined,linesnumbered]{algorithm2e}
\usepackage{setspace}
\usepackage{macros}
\usepackage[american]{babel}
\usepackage{csquotes}
\usepackage[style=apa,backend=biber,natbib=true]{biblatex}
\usepackage[hidelinks]{hyperref}
\ifdefined\pdfsuppressptexinfo\pdfsuppressptexinfo=-1\fi
\ifdefined\pdfinfoomitdate\pdfinfoomitdate=1\fi

\graphicspath{{figures/}}

\journal{Expert Systems with Applications}

\begin{document}
\csname nolinenumbers\endcsname

\begin{frontmatter}

\title{Learning to bid and bundle: a posterior-sampling agent for online retail}

\author[uu]{Hadi Mohammadi\corref{cor1}}
\ead{h.mohammadi@uu.nl}
\cortext[cor1]{Corresponding author.}
\address[uu]{Department of Methodology, Statistics \& Data Science,
Utrecht University, Padualaan 14, 3584 CH Utrecht, The Netherlands}

\begin{abstract}
An online retailer launching a perishable product bids for
sponsored-search clicks, discounts a bundle with an always-stocked staple
and displays its remaining stock, which raises willingness to pay. It
knows neither how clicks respond to its bid nor how strong and dispersed
its market is, and learns both over repeated seasons from data its own
decisions censor: a zero bid reveals nothing about paid clicks, and
choices follow only clicks. We design a posterior-sampling agent for it.
Each period it draws a market from an exact posterior and posts the bid
and discount of that market's exactly solved season plan, so every action
is one a fully informed seller could take. In a simulated online florist
it has the lowest regret averaged over markets. In weak markets it gives
up 12\% to 23\% of one season's revenue over 500 seasons, four to seven
times less than drawing once per season. A planner acting on point
estimates can stop bidding and never learn that bidding pays. The agent
escapes after about three seasons on average, against fifty with one draw
per season. Where displayed scarcity keeps willingness to pay above the
posted price, the discount needs no learning. Customers waiting for the
discount to return to its recent level look like lost sales, so frequent
waiting makes the agent underrate its market. One discount per season
removes waiting at a cost of 5.9\% of clairvoyant revenue in a weak market
and none in a strong one.

\end{abstract}

\begin{keyword}
Online retailing \sep Sponsored search advertising \sep Bundle pricing \sep
Dynamic pricing \sep Thompson sampling \sep Reinforcement learning \sep
Scarcity effect
\end{keyword}

\end{frontmatter}
\csname linenumbers\endcsname
\hypersetup{pdfauthor={Hadi Mohammadi}} 

\section{Introduction}
\label{sec:intro}

Online retailers buy traffic through sponsored search, offer a second item
at a discount on the product page, and show shoppers how many units are
left. Each lever matters on its own: paid listings partly take traffic
that organic results would otherwise bring \citep{Moshary2025}, scarcity
cues raise booking intentions on hospitality platforms, partly through
perceived value \citep{Teubner2020}, and when to disclose limited
inventory is itself a decision that shapes sales \citep{Liu2026}. An online
florist with fresh-cut bouquets to sell before the week is out uses all
three. Each period it bids for the search keyword that brings shoppers to
its page and sets a discount on a bouquet bought with a glass vase, while
the page shows how many bouquets are left. With demand known, the bid and
the discount can be planned together by dynamic programming
\citep{Ye2015,Tunuguntla2019,Mohammadi2026}. At launch the florist knows
little. It does not know how strongly clicks respond to its bid, how much
its shoppers will pay, or how dispersed they are, and it has to learn all
three from the clicks and sales that its own bids and discounts generate.

Two features make this harder than learning a posted price. The
instruments censor the data. Organic visitors keep arriving without a bid,
but only a positive bid reveals how paid clicks respond, and a purchase
choice is seen only after a click. A seller that concludes advertising
does not pay and stops bidding will never see evidence to the contrary.
The discount, in turn, reaches customers who can wait: when it moves
within the season, patient shoppers hold out for a better one, and in the
seller's data a waiting shopper looks like a lost sale. Displayed scarcity
adds a twist of its own. Deep scarcity lifts every shopper's willingness
to pay above the posted price, so low-stock periods say little about the
market.

We design and test a decision agent for this setting, with an offline
planning module and an online learning loop.
Offline, it solves the season program exactly by backward induction for
every candidate market and click curve on a grid and stores the plans.
Online, it keeps an exact posterior over the candidates, which splits into
a click part and a choice part, and in each period draws a candidate and
posts the bid and discount that the candidate's plan prescribes at the
current stock and time. A decision is a table lookup, and every action is
one a fully informed seller would take in some market the data still
allow. Drawing once per season instead gives posterior sampling for
reinforcement learning \citep{Osband2013}, which is easier to analyze and
underlies a variant that holds one discount for the whole season. We test
both in a seeded simulation of an online florist against certainty
equivalence, randomized and scripted exploration, model-free learners, and
oracles that know part of the truth.

Four findings stand out. Drawing every period is the version to deploy.
It has the lowest regret averaged over markets (\$33.6 over 500 seasons,
against \$74.0 for certainty equivalence and \$109.4 for one draw per
season), and in weak markets it gives up 12\% to 23\% of one season's
revenue over 500 seasons, four to seven times less than drawing once per
season. Certainty equivalence does slightly better only where its own
plans keep producing informative data, as in the strong market (\$11
against \$25), and it is the policy that can stop bidding for good.
Tabular Q-learning, which needs no demand model, still earns 9\% less than
the oracle after 500 seasons in the strong market. Under a
pessimistic prior that trap occurs, and posterior sampling escapes it at a
rate we can predict. With 2\% prior weight on responsive click curves,
certainty equivalence never bids and forgoes more than half of the
clairvoyant revenue every season, while one draw per season first bids
after fifty seasons on average and one draw per period after three.
Displayed scarcity decides what must be learned. Where it keeps every
shopper's willingness to pay above the posted price, the optimal discount
is a harvest level that depends only on known quantities, and the only
market question left is whether the market is of that kind. Customers who
wait contaminate learning. Waited bundles are recorded as no-sales, so
once waiting is frequent the agent's beliefs drift toward weaker markets,
whichever schedule it uses. Holding one discount for the season removes
waiting at a cost of 5.9\% of clairvoyant revenue in a weak market and none
in a strong one.

The paper adds to work on learning agents for e-commerce decisions in
design, testing and guidance. On design, the agent learns a per-click bid
and a bundle discount together from data censored twice, with an exact
posterior and an offline library of exactly solved plans, so its actions
stay readable and its computation small. On testing, it is compared with
standard and model-free learners in strong and weak markets and under two
stress tests: a pessimistic prior that traps certainty equivalence, for
which we derive how fast posterior sampling escapes under each sampling
schedule (Proposition~\ref{prop:trap}), and customers who wait for a
better discount, whose waits bias the agent once they are frequent
(Lemma~\ref{lem:contamination} gives the threshold for a frozen plan).
The analysis also shows that displayed scarcity can make discount learning
unnecessary (Proposition~\ref{prop:saturated}). On guidance,
Section~\ref{sec:managerial} turns the results into deployment rules with
the signals a seller can watch.

Two unpublished manuscripts are closest to this one.
\citet{Mohammadi2026} solves the season model used here with demand known.
We restate that model in full, so the paper can be read on its own, with
one change, and study it under learning, which that work does not address.
\citet{Mohammadi2026b} studies a seller with unlimited inventory that
learns which of a few prices to post, with no bid or bundle, against
customers who wait. Section~\ref{sec:strategic} applies its
buy-wait-leave rule, which it takes from \citet{Nabi2018}, to the bundle
discount. Section~\ref{sec:literature} sets out what each contributes
here.

Section~\ref{sec:literature} reviews related work.
Sections~\ref{sec:model} to~\ref{sec:theory} present the model, the agent
with its computing cost, and the analytical results,
Sections~\ref{sec:results} and~\ref{sec:strategic} the experiments, and
Section~\ref{sec:managerial} the rules for deployment.
Section~\ref{sec:conclusion} concludes. Proofs are in \ref{app:proofs}.
The online supplement holds numerical checks and additional exhibits but
no proofs.

\section{Related literature}
\label{sec:literature}

\paragraph{Learning agents for pricing}
Reinforcement learning is a common tool for pricing in intelligent-systems
research. Q-learning has priced interdependent perishable products over a
finite horizon \citep{RanaOliveira2015}, deep reinforcement learning has
set prices jointly with inventory under reference-price effects
\citep{Zhou2022} and for perishable products that spoil
\citep{Kavoosi2025}, multi-agent learning has priced several perishable
products at once \citep{Qiao2024}, and policy-gradient methods have priced
against online reviews \citep{Wu2023}. These agents learn a value
function or a policy from many simulated episodes and need no demand
model. A multi-armed bandit with Thompson sampling and upper confidence
bounds prices a delivery firm's services from real data
\citep{Tajik2024}, and reinforcement learning allocates advertising
budgets across channels \citep{WangZareeihemat2025}. Our agent is
model-based instead: it learns three demand parameters and plans exactly,
which keeps its actions readable and its data needs small. In operations
research, learning while pricing a fixed stock over a finite season
\citep{Gallego1994} is well studied
\citep{AvivPazgal2005,Araman2009,Besbes2009,denBoer2015}, and with
repeated selling seasons a slightly modified certainty-equivalent policy
learns on its own, because optimal prices vary with the remaining stock
\citep{denBoerZwart2015}. Learning while pricing extends to general choice
models \citep{Broder2012}, perishable ordering \citep{Keskin2022}, priors
shared across products \citep{Bastani2022}, add-on discounts
\citep{SimchiLevi2022} and Thompson-sampling bundling of two products
\citep{Fang2025}, none with an advertising lever or scarcity-dependent
willingness to pay. For perishable products without replenishment,
\citet{Yang2022} learn the willingness-to-pay distribution from
early-season sales with a neural network, and for e-commerce promotions
without price-elasticity history, \citet{GreensteinMessica2020} pair a
demand model with gradient boosting. Thompson sampling entered
operational pricing with \citet{Ferreira2018} and is analyzed by
\citet{Osband2013}, \citet{Russo2014} and \citet{AgrawalGoyal2017}.
Closest in method, \citet{Shimizu2024} apply posterior sampling to an
episodic revenue-management problem whose prices come from a linear
program re-solved each episode. Learning can stop short of the truth even
under optimal experimentation \citep{Rothschild1974}, and does so when
the seller acts on its estimates
\citep{Harrison2012,Keskin2014,denBoer2014}, which
\citet{Huh2022} remedy by Bayesian dithering, finite parameter sets
included. In our model the trap reappears on the advertising side, where
the instrument that generates information is bought per click.

\paragraph{Advertising and pricing}
With demand known, \citet{Ye2015} optimize a sponsored-search bid jointly
with a posted price over a selling season, \citet{Tunuguntla2019} add
inventory-linked willingness to pay for a perishable product, and
\citet{Salehi2023} a limited advertising budget. \citet{Agrawal2025} set
campaign prices and exchange bids in one exactly solved Markov decision
process on an advertising agency's side, with no retail price, goods
inventory or bundle, and \citet{LiYang2022} feed Bayesian click-through
estimates from real campaign logs into a static keyword-targeting model.
\citet{Mohammadi2026} replaces the posted price with a bundle discount
under scarcity-dependent reservation prices and establishes the
bid--discount separation and complementarity that our agent's plans
inherit. With learning, the nearest published counterpart is
\citet{Gurkan2026}, who learn demand and advertising response while
setting price, advertising level and inventory, giving sufficient
conditions for fast learning and an asymptotically optimal policy. Their
advertising decision is a level, ours a per-click auction bid whose zero
level leaves the click data uninformative about the paid click response. \citet{LiuSun2026} price while learning an unknown advertising
model from a class that includes the Nerlove--Arrow and Vidale--Wolfe
models and show that static policies are near-optimal for the classic
models. Online learning in repeated auctions studies bidders who learn
values or competition \citep{Weed2016,Balseiro2019,Han2025,Guo2026,Yang2025},
and auto-bidding agents set each auction's bid against value or
return-on-spend targets \citep{Aggarwal2024}. In all of these the bid is
the only lever and inventory is absent. Here the bid interacts with a
pricing lever through a perishable stock, and what the bid censors is the
seller's own demand data: a zero bid hides the click curve.

\paragraph{Bundling and displayed scarcity}
Bundling and cross-selling models price a second product against limited
inventories without an advertising decision
\citep{Netessine2006,Bulut2009,Gokgur2019,Song2021}. With
name-your-own-price bids and an uncertain supplier cost, complementarity
and marginal cost decide whether the bundle beats separate sale
\citep{Hemmati2023}, and \citet{Abdallah2026} show that as inventory
becomes scarce relative to market size, optimal dynamic mixed bundling
converges to component pricing. The harvest discount of
\citet{Mohammadi2026} points the same way: under deep scarcity the
discount only prices the abundant staple into the bundle. We read that
result for learning. Where displayed stock keeps willingness to pay above
the posted price, the discount needs no learning.

\paragraph{Customers who wait and learning sellers}
Forward-looking customers who time purchases against markdowns are
classic \citep{Aviv2008,Su2008,Cachon2009}. When they know the stockout
risk, a larger initial stock sends more of them to wait \citep{Zhao2022}.
Prices that move in one direction only are a known remedy
\citep{Chen2024}, and e-commerce platforms choose when coupons can be
redeemed with strategic consumers in mind \citep{Li2020}. For a learning
seller, \citet{AvivWei2019} find that committing to a price increasingly
dominates responsive pricing as the share of strategic consumers grows.
Near-optimal markdown policies for demand learning with forward-looking
customers delay markdowns \citep{Birge2025}, \citet{ZhangJasin2022} learn
valuation and patience and converge to cyclic pricing, and customers also
react to experimentation visible in a learning seller's prices
\citep{Golrezaei2021,ChenGao2022}. In \citet{Mohammadi2026b} a seller with
unlimited inventory learns which of a few prices to post against
patience-driven waiting. Its benchmark posts one price throughout, so
waiting there comes only from the learner's own price changes. We apply
the buy-wait-leave rule it takes from \citet{Nabi2018} to the bundle
discount of a perishable stock, where the clairvoyant discount itself
moves within the season, and show that waits recorded as no-sales bias a
myopic likelihood, a misspecification of the kind studied by
\citet{Berk1966} and, in revenue management, by \citet{Cooper2006}.

\begin{table}[!t]
\centering
\caption{How the closest learning studies differ from this paper.}
\label{tab:positioning}
\footnotesize
\setlength{\tabcolsep}{3pt}
\newcolumntype{L}[1]{>{\raggedright\arraybackslash}p{#1}}
\begin{tabular}{@{}L{2.6cm}L{3.2cm}L{3.0cm}L{3.9cm}L{2.4cm}@{}}
\toprule
Study & Decisions & Learned & Feedback & Customers \\
\midrule
\citet{RanaOliveira2015}; \citet{Qiao2024} & prices of perishable
products & a policy or value function, model-free & sales & myopic \\
\citet{denBoerZwart2015} & price of a finite stock over repeated seasons
& demand parameters & sales & myopic \\
\citet{Ferreira2018} & prices against finite stocks & demand parameters &
sales & myopic \\
\citet{Gurkan2026} & price, advertising level, inventory & demand and
advertising response & sales & myopic \\
\citet{ZhangJasin2022} & prices & valuations and patience & purchases
over time & patient \\
This paper & per-click bid and bundle discount for a fixed stock & click
curve, market strength and spread & clicks, informative only at positive
bids; choices only after a click & myopic, or waiting for a better
discount \\
\bottomrule
\end{tabular}
\end{table}

Table~\ref{tab:positioning} sets the paper apart on what is learned and
on what the data reveal. Model-free agents need no demand model but many
episodes. Model-based learners in operations research learn a few
parameters from sales that every posted price reveals. Here the instrument
that buys information is a per-click bid whose zero level reveals nothing,
choices arrive only after a click, and scarcity and waiting change what
the choices say about the market. The results of the paper are about
these features: the trap and how fast sampling escapes it, what scarcity
removes from the learning task, and how waiting biases the learner.

\section{Model and learning environment}
\label{sec:model}

\subsection{The selling season}
\label{sec:season}

We adopt the season model of \citet{Mohammadi2026} and state it here in
full, in our notation and with one change: we anchor its click curve at the
organic rate. An online retailer sells two commodities over a
season of $T$ periods, indexed by the number of periods remaining
$t = T, \dots, 1$. The \emph{primary} commodity is perishable: the season
starts with $I_0$ units, there is no replenishment, and units left at the
end have salvage value $s$ per unit (baseline $s = 0$). The \emph{secondary}
commodity is an abundant staple that never stocks out. Component prices
$p_P$ and $p_S$ are fixed for the season. The two products are also offered
as a bundle at price $p_P + p_S - d_t$, where the \emph{bundle discount}
$d_t \in [0, d_{\max}]$ is one of the retailer's two per-period controls.

The other control is a cost-per-click \emph{sponsored-search bid} $b_t \in
[0, b_{\max}]$. Periods are short enough that at most one customer clicks per period;
a click arrives with probability $\ctr{b}$ and costs $b$ whether or not a
purchase follows. We use the S-shaped click curve
\begin{equation}
\label{eq:ctr}
\ctr{b} \;=\; \lambda_0 \;+\; (\lambda_\infty - \lambda_0)\,
\frac{\sigma(\ctrslope b - \ctroffset) - \sigma(-\ctroffset)}
     {1 - \sigma(-\ctroffset)},
\qquad \sigma(z) = \frac{1}{1 + e^{-z}},
\end{equation}
anchored so that $\lambda(0) = \lambda_0$ exactly for every location
$\ctroffset$: without a bid the retailer gets its organic click rate
$\lambda_0$ and learns nothing about the paid part of the curve. (The
curve of \citet{Mohammadi2026} is not anchored, and Section~S1 of the
online supplement re-solves the oracle under it.) A positive floor matches the
evidence that paid listings partly draw traffic that organic listings
would otherwise receive \citep{Moshary2025}, so a seller that stops
bidding keeps selling and never learns why. Every click is billed at the
bid, organic ones included. This overstates the cost of bidding, a
limitation Section~\ref{sec:conclusion} returns to. The slope $\ctrslope$ and the floor and cap
$\lambda_0, \lambda_\infty$ describe the retailer's own listing and are
known from history. The location $\ctroffset$ reflects the competitiveness of the
keyword auction and is unknown for a new product. Competing advertisers
enter through it, so a single seller learning its click curve covers
auction competition in reduced form.

A clicking customer sees the posted discount and the remaining inventory
$I$, and displayed scarcity shifts willingness to pay (WTP): the reservation price
for the primary commodity is
\begin{equation}
\label{eq:wtp}
R_P(I) \;=\; \mu_P(I) + \sigma_P(I)\,X, \qquad
\mu_P(I) = \delta_\mu e^{-k I}, \qquad
\sigma_P(I) = 1 + \delta_\sigma e^{-k I},
\end{equation}
with $X \sim \mathrm{Gamma}(K_P, \scP)$, while $R_S \sim
\mathrm{Gamma}(K_S, \scS)$ is inventory-independent and the bundle
reservation price is additive, $R_B(I) = R_P(I) + R_S$
\citep{Adams1976}; $F_P(\cdot \mid I; \parmw)$ and $F_S$ denote the distribution
functions of $R_P(I)$ and $R_S$. The scalers $\delta_\mu$ and $\delta_\sigma$ summarize
the market's \emph{strength} and \emph{dispersion} and are unknown for a
new product. The display sensitivity $k$, the Gamma shapes $K_P, K_S$
and the scales $\scP, \scS$ are known. The customer buys the
surplus-maximizing option among the primary alone ($R_P(I) - p_P$), the secondary alone
($R_S - p_S$), and the bundle ($R_B(I) - p_P - p_S + d$), provided its
surplus is positive; $\aP, \aS, \aB, \aN$ denote the resulting choice
probabilities at $(I, d)$, the last for no purchase. Customers in the
main model are myopic. Section~\ref{sec:strategic} lets them wait for a
better discount.

\subsection{The full-information oracle}
\label{sec:oracle}

Given the unknown primitives $\parm$ (made precise below), let
$\Vfun{I}{t}$ be the optimal expected remaining revenue with $t$ periods
and $I$ units left, and $\marg{I}{t} = \Vfun{I}{t-1} - \Vfun{I-1}{t-1}$
the marginal value of the $I$-th unit. The Bellman recursion is
\begin{align}
\label{eq:bellman}
\Vfun{I}{t} \;&=\; \Vfun{I}{t-1} \;+\;
\max_{b \le b_{\max},\, d \le d_{\max}}\; \ctr{b}
\Bigl[\, \inner_I\bigl(d; \marg{I}{t}, \parm\bigr) - b \,\Bigr],
\\
\inner_I(d; \Delta, \parm) \;&=\; (\aP + \aB)\,(p_P - \Delta)
\;+\; (\aS + \aB)\,p_S \;-\; \aB\, d,
\notag
\end{align}
with $\Vfun{0}{t} = 0$, $\Vfun{I}{0} = sI$, and the choice probabilities
in the inner objective $\inner_I$ evaluated at $(I, d; \parmw)$, so the
state enters through the scarcity level and through the marginal value.
For \emph{every} parameter value, $\inner_I$ does not depend on the bid,
so each period's maximization splits into two one-dimensional searches,
and both optimal controls are nonincreasing in $\marg{I}{t}$
\citep{Mohammadi2026,Topkis1998}: where stock is scarce relative to
remaining time, $\Delta$ is high and paid traffic cannot pay for itself.
Numerically this gives an on--off bid with a \emph{switching curve}
$\swc{t}$ (the lowest inventory at which the bid is on with $t$ periods
remaining) and, in strong markets, a discount pinned at the \emph{harvest
level} $d_H = \arg\max_d\, \harv(d)$ with $\harv(d) = (p_S -
d)\bigl(1 - F_S(p_S - d)\bigr)$, which prices the secondary commodity
into the bundle (Section~\ref{sec:theory} makes this limit precise).

\subsection{The learning problem}
\label{sec:learning}

The retailer knows the model structure but not the demand primitives,
which we collect in $\parm = (\parmc, \parmw)$ with $\parmc = \ctroffset$
and $\parmw = (\delta_\mu, \delta_\sigma)$. We let $\parm$ range over a
finite grid $\parmspace = \parmspace_c \times \parmspace_w$ of candidate
values with $J = |\parmspace|$. The true parameter is $\parmtrue$. The
grid spans weak to strong markets and cheap to expensive click auctions,
so learning means discovering which market regime the product lives in,
the question a merchandiser faces at launch.

Selling proceeds over repeated seasons $n = 1, \dots, \nseasons$
(fashion-like assortment cycles): each season resets inventory to $I_0$ and
runs $T$ periods under the same unknown $\parmtrue$. Within a period at
state $(t, I)$ the retailer posts $(b, d)$ and observes two signals:
\begin{equation}
\label{eq:obs}
c \sim \mathrm{Bernoulli}\bigl(\ctrtheta{b}{\parmc}\bigr),
\;\text{and, only if } c = 1,\;
y \in \{P, S, B, \emptyset\} \ \text{with probabilities }
\alpha_{y}(I, d; \parmw).
\end{equation}
Clicks are billed, hence observed. Item-level sales identify $P$, $S$, and
$B$, and a billed click without a sale identifies $\emptyset$
(Figure~\ref{fig:environment}). The choice
signal is \emph{censored by the click}: periods without a click carry no
information about $\parmw$ and, by the anchoring in \eqref{eq:ctr},
periods without a bid carry none about $\parmc$. A season ends at $t = 0$
or at stockout, so successful policies self-censor late-season choice
data.

Table~S.1 in the online supplement collects the notation.
The retailer starts from a prior $\post{0}$ on $\parmspace$ and updates to
$\post{n}$ after season $n$. Performance is measured against the
full-information oracle: with $\Vstar(\parm) = \Vfun{I_0}{T}$ the oracle
season value and $V_{\parmtrue}(\pi)$ the exact expected season revenue of
a policy $\pi$ under the true parameters, per-season regret and cumulative
Bayes regret are
\begin{equation}
\label{eq:regret}
r_n \;=\; \Vstar(\parmtrue) - V_{\parmtrue}(\pi_n),
\qquad
\regret{\nseasons} \;=\;
\E\Bigl[\, \textstyle\sum_{n=1}^{\nseasons} r_n \,\Bigr],
\end{equation}
where $\pi_n$ is the policy deployed in season $n$ and the expectation is
over the prior, the observation noise, and the algorithm's randomization.
For a learner that plays one season plan, $V_{\parmtrue}(\pi_n)$ is exact
after one backward pass, so its regret estimates carry no simulation noise
beyond which plan was deployed. A learner whose actions respond to data
within the season, the per-period agent of Section~\ref{sec:algorithms},
$\varepsilon$-greedy and Q-learning (Section~\ref{sec:benchmarks}), plays
no fixed plan. Its regret is estimated as the oracle's realized revenue minus its
own, with the same customers facing both (common random numbers), which is
unbiased.

\section{The learning agent}
\label{sec:algorithms}

\begin{figure}[!t]
\centering
\includegraphics[width=0.96\textwidth]{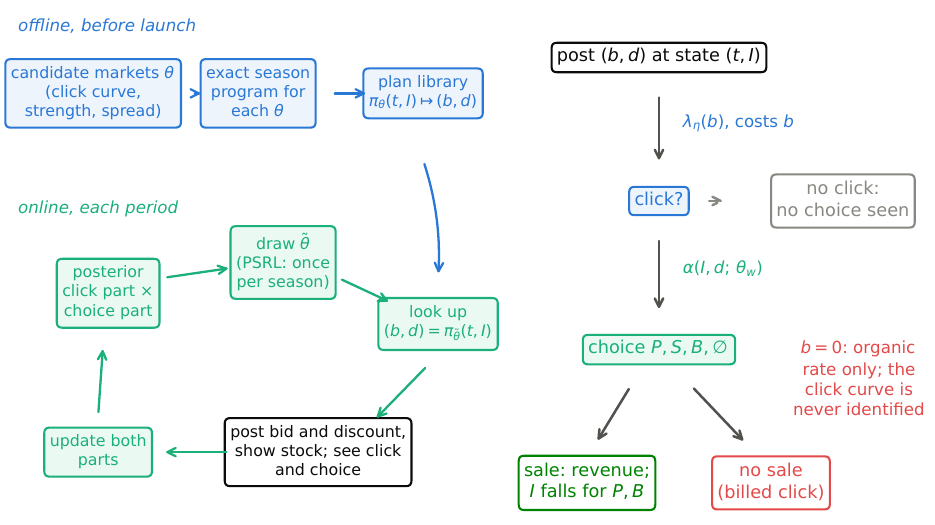}
\caption{The learning agent (left) and the events within a period (right);
solid outcomes are observed.}
\label{fig:environment}
\end{figure}

Figure~\ref{fig:environment} shows the agent's two parts. Offline, before
the first season, it solves the season program \eqref{eq:bellman} for every
candidate in $\parmspace$ and stores each plan as a bid map and a discount
map over the remaining time and stock. Online, it keeps the posterior as
two vectors, one over click curves and one over willingness-to-pay
parameters, updates them after every click and choice, and in each period
draws a candidate and looks up that candidate's action.

\subsection{Likelihood and exact posterior}
\label{sec:posterior}

Write $\hist{n}$ for the history through season $n$: the posted controls,
click outcomes, and post-click choices at their visited states. The
likelihood factors into a click part and a choice part,
\begin{equation}
\label{eq:factorization}
L(\hist{n} \mid \parm) \;=\;
\underbrace{\prod_{\text{periods}}
\ctrtheta{b}{\parmc}^{\,c}\,\bigl(1 - \ctrtheta{b}{\parmc}\bigr)^{1-c}}
_{\text{depends only on } \parmc}
\;\times\;
\underbrace{\prod_{\text{clicks}} \alpha_{y}(I, d; \parmw)}
_{\text{depends only on } \parmw},
\end{equation}
because the click depends on the bid alone and the choice, given a click,
on $(I, d)$ alone. The posted controls depend on $\parm$ only through
past data, so they add no factor that involves $\parm$. A product prior
therefore stays a product under any policy: clicks update only the
$\parmc$-marginal and choices only the $\parmw$-marginal. The agent keeps
the posterior exactly as two vectors on $\parmspace_c$ and
$\parmspace_w$, with one log-likelihood update per observation. Draws,
means and modes are exact up to a $10^{-12}$ floor on choice likelihoods
for numerical stability (Section~S1 of the online supplement).

\subsection{The agent and its benchmarks}
\label{sec:psrl}
\label{sec:benchmarks}

The plans $\pi_\parm$ are solved by backward induction with ties broken
toward the smallest control. With the discount separated from the bid and
the choice tables precomputed, each plan takes milliseconds
(Section~\ref{sec:cost}). Algorithm~\ref{alg:psrl} then draws a candidate
from the posterior in every period and posts the action that the
candidate's plan prescribes at the current state. We call this \textsc{ts-update}.
Like the Thompson-sampling algorithms of \citet{Ferreira2018}, it samples
every period, and the stored plans already respond to the remaining stock
and time. Drawing only at the start of each season and playing that plan
to the end is posterior sampling for reinforcement learning
(\textsc{psrl}) of \citet{Osband2013}, the form that
Proposition~\ref{thm:regret} covers and that Section~\ref{sec:strategic}
restricts to one discount per season. Either way every action is one that
the full-information plan of some candidate prescribes, which at our
calibration means a bid that switches on and off along the plan's
switching curve and a harvest or clearance discount
(Section~\ref{sec:oracle}). \textsc{psrl}
follows one plan for a whole season; \textsc{ts-update} may switch plans
within the season as the data arrive, which is what lets it act on early
evidence.

\begin{algorithm}[t]
\caption{The learning agent (\textsc{ts-update}; the boxed line is
skipped by \textsc{psrl}, which draws only at season start)}
\label{alg:psrl}
\KwIn{grid $\parmspace$, prior $\post{0}$, stored plans
$\{\pi_\parm\}_{\parm \in \parmspace}$}
\For{season $n = 1, \dots, \nseasons$}{
  draw $\parmsamp{} \sim$ current posterior; set $I \leftarrow I_0$\;
  \For{$t = T, \dots, 1$ \textbf{while} $I > 0$}{
    post $(b, d) = \pi_{\parmsamp{}}(t, I)$; observe click $c$ and, if
    $c = 1$, the choice $y$; collect revenue\;
    update the click posterior with $(b, c)$ and, if $c = 1$, the choice
    posterior with $(I, d, y)$\;
    \fbox{re-draw $\parmsamp{} \sim$ current posterior}\;
  }
}
\end{algorithm}

\paragraph{Benchmarks}
We compare against the \emph{clairvoyant oracle} $\pi_{\parmtrue}$;
\emph{certainty equivalence} (\textsc{ce}), which solves
\eqref{eq:bellman} each season for the posterior-mean click curve and
choice tables; $\varepsilon$-\emph{greedy}, \textsc{ce} with a decaying
per-period probability of a uniform random control pair;
\emph{explore-then-commit}, a scripted control sweep for a fixed number of
seasons followed by \textsc{ce}; \emph{static-pair Thompson sampling}, a
model-free yardstick that treats each constant $(b, d)$ pair on a coarse
grid as an arm with the season return as reward; the \emph{best static
pair} under the true parameters, the strongest clairvoyant benchmark that
does not adjust within the season; two \emph{half oracles} that know one
block of $\parmtrue$ and learn the other; and \emph{tabular Q-learning}, a
model-free reinforcement learner in the manner of \citet{RanaOliveira2015}
that learns action values over the states $(t, I)$ and 36 fixed
bid--discount pairs from the revenue it earns, exploring
$\varepsilon$-greedily.

\subsection{Computing cost}
\label{sec:cost}

Table~\ref{tab:cost} reports what the agent costs to run, measured in
Python with one process on an Apple M4 Max. Most of the offline time goes
into the choice tables, about $0.17$ seconds per willingness-to-pay
candidate. Once they exist, a season plan takes about $0.26$ milliseconds,
so the $125$ plans of the main grid take $0.03$ seconds and $0.16$ MiB.
Online, a decision (one posterior draw and one table lookup) takes about
$3$ microseconds and an update after an observation under one, whatever
the grid size. The agent never solves anything while it sells.

\begin{table}[!t]
\centering
\caption{Computing cost of the agent by grid size $J$: choice tables and
all season plans (offline), plan memory, and time per decision and per
update (online).}
\label{tab:cost}
\small
\begin{tabular}{rrrrrr}
\toprule
 & \multicolumn{3}{c}{Offline} & \multicolumn{2}{c}{Online} \\
\cmidrule(lr){2-4}\cmidrule(lr){5-6}
$J$ & Tables (s) & Plans (s) & MiB & Decision ($\mu$s) & Update ($\mu$s) \\
\midrule
27 & 1.5 & 0.007 & 0.03 & 2.7 & 0.64 \\
125 & 4.2 & 0.033 & 0.16 & 2.9 & 0.63 \\
405 & 13.6 & 0.106 & 0.52 & 3.7 & 0.68 \\
891 & 16.6 & 0.237 & 1.14 & 4.1 & 0.70 \\
\bottomrule
\end{tabular}
\end{table}

\section{Design properties}
\label{sec:theory}

This section states what the agent can rely on: which data inform which
part of the posterior, what displayed scarcity removes from the task, how
fast sampling leaves a pessimistic prior, and the regret guarantee of the
once-per-season form. Proofs are in \ref{app:proofs}. Throughout,
$\parmspace$ is finite and a season's revenue lies in an interval of length
$\bar{V} = (p_P + p_S + b_{\max})\,T + s I_0$: each period earns at most
$p_P + p_S$ and pays at most $b_{\max}$ for a click, and salvage adds at
most $s I_0$.

\subsection{What scarcity removes from the learning task}

Where scarcity carries the price, the discount can be frozen at a level
that needs no learning:

\begin{proposition}[Harvest limit: strong markets need no discount
learning]
\label{prop:harvest}
Fix $I$ and let $\varepsilon = F_P(p_P \mid I)$, the probability that the
primary reservation price fails to clear the undiscounted primary price.
Then, uniformly over $d$ and over compact $\Delta$-sets,
\begin{gather*}
\bigl|\, \inner_I(d; \Delta, \parm) - \bigl[(p_P - \Delta) + \harv(d)
\bigr] \,\bigr| \;\le\; 2\,\varepsilon\,\bigl(|p_P - \Delta| + 2 p_S +
d_{\max}\bigr), \\
\harv(d) = (p_S - d)\bigl(1 - F_S(p_S - d)\bigr),
\end{gather*}
and $\harv$ depends only on the \emph{known} secondary taste
distribution. As $\varepsilon \to 0$ the optimal discount converges to
the harvest level $d_H = \arg\max_d \harv(d)$, independent of $\Delta$,
$I$, $t$, and $\parmw$. (At the baseline calibration the residual is
below $10^{-13}$ and $d_H = 5.4$ with $\harv(d_H) = 10.17$.)
\end{proposition}

With $\varepsilon = 0$ at every reachable inventory level this is the
harvest discount of \citet{Mohammadi2026}. The bound carries it to
markets that are only nearly saturated. When scarcity keeps willingness
to pay high, the discount only prices the secondary commodity into the
bundle, and that trade-off involves nothing the retailer needs to learn.

The bid is the agent's only source of click-curve data, and the stock
extremes say little about the market:

\begin{lemma}[Censoring and saturation]
\label{lem:censoring}
(i) A policy that never bids generates data whose likelihood is constant in
$\parmc$, so under a product prior the click-curve posterior never moves. (ii) Conditional on the
secondary draw, the choice is a piecewise-constant function of $R_P$ whose
decisive thresholds all lie in the window $W(d) = [\,p_P - d,\
p_P\,]$. Hence, for any two $\parmw, \parmw'$, the
total-variation distance of the choice laws satisfies
\[
\TV{\alpha(I, d;\, \parmw)}{\alpha(I, d;\, \parmw')}
\;\le\; 2 \sup_{x \in W(d)}
\bigl|\, F_P(x \mid I;\, \parmw) - F_P(x \mid I;\, \parmw') \,\bigr|.
\]
In particular, if both reservation-price distributions place at most
$\varepsilon$ of their mass below the top of the window (deep scarcity),
or both at most $\varepsilon$ above its bottom (dead market), the choice
laws differ by at most $2\varepsilon$: both inventory extremes are
uninformative about $\parmw$.
\end{lemma}

Part (i) is exact because of the anchored click curve \eqref{eq:ctr}.
Part (ii) is visible at the calibration of Section~\ref{sec:setup}: the
choice laws of the strongest and weakest markets are
indistinguishable at low inventory, where scarcity lifts every market's
willingness to pay above the window, and separate almost completely at
season-start inventory ($F_P(p_P \mid I)$ is below $10^{-5}$ for both
at $I \le 2$, against $0.00$ and $1.00$ at $I = 15$).

Season-start states are informative, and every new season revisits them,
so learning is fast.

In strong markets the willingness-to-pay side reduces to a single
yes-or-no question:

\begin{proposition}[Saturated markets: one class, one policy, one bit]
\label{prop:saturated}
Call $\parmw$ \emph{saturated} if $F_P(p_P \mid I; \parmw) = 0$ at every
inventory level the season can reach; since $X \ge 0$ in \eqref{eq:wtp},
$\mu_P(I)$ is the smallest value $R_P(I)$ can take and the condition is
the closed-form one $\delta_\mu e^{-k I_0} \ge p_P$, and let $\parmspace_{\mathrm{sat}} \subseteq \parmspace_w$
collect the saturated grid points. Then
(i) for every saturated $\parmw$ and every $(I, d)$, $\aS = \aN = 0$ and
$\aP = F_S(p_S - d) = 1 - \aB$: all saturated parameters share a single
choice law, free of $\parmw$ and of $I$;
(ii) their season programs coincide and reduce to
$\Vfun{I}{t} = \Vfun{I}{t-1} + \max_{b, d} \ctr{b}\bigl[(p_P -
\marg{I}{t}) + \harv(d) - b\bigr]$, so value, optimal bid and optimal
discount are common to all of them, and the optimal discount is the
harvest discount $d_H$ in every state;
(iii) hence the choice likelihood is constant on
$\parmspace_{\mathrm{sat}}$: under any policy and any product prior, the
$\parmw$-posterior restricted to $\parmspace_{\mathrm{sat}}$ is the prior restricted to
$\parmspace_{\mathrm{sat}}$ for every $n$, and in a season in which the
sampled and the true WTP blocks are both saturated the regret is exactly
that of the click-curve sample;
(iv) the boundary is identified one-sidedly: a single no-purchase or
secondary-only purchase has probability zero under every saturated
parameter and eliminates $\parmspace_{\mathrm{sat}}$ at once, while a
saturated truth never generates one, so the class separates from the rest
of the grid only at the rate of its Kullback--Leibler ($\mathrm{KL}$) divergence.
\end{proposition}

The condition, the choice law of part (i), and the reduced season program
and harvest discount of part (ii) come from the full-information analysis
of \citet{Mohammadi2026}, where the condition is an assumption. Parts (iii)
and (iv) give them their learning meaning. For a given click curve the
saturated points share one season plan, so the seller has no need to tell
them apart: on the willingness-to-pay side the question is the single bit
``is the market saturated'' (Section~\ref{sec:recovery}).

\subsection{The no-bidding trap and how fast sampling escapes it}

The sampling schedule sets how fast the agent leaves a pessimistic prior.

\begin{proposition}[The no-bidding trap]
\label{prop:trap}
Let the prior be a product prior, fix a set of click curves, called
\emph{responsive}, with prior mass $m$, and suppose (a) the posterior-mean
model's optimal bid is zero in every state, along every posterior
reachable without bidding, and (b) at every state a season reaches without
a bid, a sampled plan bids if its click curve is responsive and does not
bid otherwise. Then:
(i) \textsc{ce} never bids, its click-curve posterior stays at the prior
forever, and if bidding is profitable under $\parmtrue$ it loses at least
a fixed positive amount every season: its regret grows linearly in
$\nseasons$.
(ii) \textsc{psrl} first bids in a season whose index is geometric with
success probability $m$, so the mean is $1/m$ seasons.
(iii) \textsc{ts-update} first bids in a period whose index is geometric
with success probability $m$. If seasons without a bid last $T$ periods,
its first bidding season is geometric with success probability
$1 - (1 - m)^T$, with mean $1/[1 - (1 - m)^T]$, close to $1/(Tm)$ when
$Tm$ is small.
\end{proposition}

Premise (a) can be checked by season solves and holds, for example, when
enough prior mass sits on a click curve whose midpoint lies beyond the
feasible bid range. This is the episodic, advertising-side analogue of the
uninformative-price traps of \citet{Harrison2012} and \citet{Keskin2014}:
acting on the belief that clicks are unresponsive destroys the only data
that could refute it. Parts (ii) and (iii) show why the sampling schedule
matters. Drawing every period gives the agent $T$ chances per season to
try a responsive click curve, so for small $Tm$ its expected wait is about
$T$ times shorter. One bidding period moves the click-curve posterior, and
learning proceeds from there. Section~\ref{sec:weak} quantifies the trap
at a calibration where the frozen belief costs more than half of the
oracle revenue per season. Premise (b) fails there only for a few
plans of the weakest market, which post no bid at the start of a season,
and the observed escape times match both formulas (S1, Figure~S.1).

Part (iii) also tells the seller how to set the prior. Under the premises,
and if seasons without a bid last $T$ periods, \textsc{ts-update} has
posted no bid after $K$ seasons with probability
$(1 - m)^{TK}$, so it bids within $K$ seasons with probability at least
$1 - \delta$ once the prior puts mass $m \ge 1 - \delta^{1/(TK)}$ on
responsive click curves. \textsc{psrl} needs $m \ge 1 - \delta^{1/K}$ for
the same guarantee. With $T = 20$, $K = 3$ and $\delta = 0.05$ the two
floors are $0.049$ and $0.632$.

\subsection{A regret guarantee}

\begin{proposition}[Bayes regret of \textsc{psrl}]
\label{thm:regret}
Under the protocol of Section~\ref{sec:learning} with finite $\parmspace$
of size $J$, any prior and $\parmtrue$ drawn from it, the cumulative Bayes
regret of
\textsc{psrl} satisfies
\[
\regret{\nseasons} \;\le\; \bar{V} \sqrt{\tfrac{1}{2}\, \lvert \Pi \rvert\,
H(\pi^{*})\, \nseasons} \;\le\; \bar{V} \sqrt{\tfrac{1}{2}\, J\, \nseasons \log J},
\]
where $\Pi = \{\pi_\parm : \parm \in \parmspace\}$ is the set of distinct
season plans, $\pi^{*} = \pi_{\parmtrue}$ is the optimal one, and $H(\pi^{*})
\le \log \lvert \Pi \rvert$ is its entropy under the prior.
\end{proposition}

This is the information-theoretic bound of \citet{Russo2016} with seasons
as rounds and season plans as actions, applied to the exact posterior; we
claim no new rate. It rules out freezing but forecasts little. At our
calibration the right-hand side is about $6.6 \times 10^5$ over 500
seasons, more than the passive policy's regret, while the regret
of \textsc{psrl} is $37$ at the baseline truth and $109$ averaged over the
prior, and it does not grow with the grid size (S1). Since $\lvert \Pi \rvert$
counts distinct plans, grid points that share one, such as the saturated
class of Proposition~\ref{prop:saturated} at a given click curve, leave
$\lvert \Pi \rvert$ unchanged. We know of no comparable bound for
\textsc{ts-update}, whose draws within a season are not independent
rounds. Its case rests on Proposition~\ref{prop:trap}(iii) and on the
numerical study.

\section{Experimental evaluation}
\label{sec:results}

\subsection{Test case and protocol}
\label{sec:setup}

The calibration describes an online florist and takes its season and
taste primitives from \citet{Mohammadi2026}. The perishable product
is a fresh-cut bouquet listed at \$50: it cannot be restocked once the
season has started, and bouquets still unsold at the end are worth nothing. The
staple is a glass vase listed at \$20 that is always in stock, and each
period the florist chooses a discount for buying the two together and a
bid for its search keyword. Monetary amounts throughout are in U.S.
dollars. No retailer's records informed the calibration. Component prices
are fixed for the season, so in the strong market the \$50 list price sits
far below what shoppers would pay and the bid and the discount are the
only levers.

Table~\ref{tab:design} collects the calibration and the experimental
design. Unless stated otherwise, a season has $T = 20$ periods and
$I_0 = 15$ units, the season of the two weak-market scenarios of
\citet{Mohammadi2026} (its own baseline season is shorter and smaller, $T = 15$ and $I_0 = 10$). The parameter grid is $\delta_\mu \in \{50, 60, 80,
110, 170\}$, $\delta_\sigma \in
\{2, 6, 12, 22, 42\}$, and $\ctroffset \in \{4, 4.75, 5.5, 6.25, 7\}$,
giving $J = 125$ candidates from near-dead to strong markets and cheap to
expensive click auctions. Priors are uniform unless stated. The
\emph{baseline} truth is the strong market $(\delta_\mu, \delta_\sigma;
\ctroffset) = (170, 42; 5.5)$, the \emph{low-mean} truth $(50, 6; 5.5)$,
and the \emph{low-dispersion} truth $(60, 2; 5.5)$. The saturation
condition of Proposition~\ref{prop:saturated} reads $\delta_\mu \ge p_P
e^{k I_0} = 105.85$ here, so the baseline truth is saturated together
with nine of the other twenty-four market grid points, while both weak truths
are not.
Under the anchored
click curve \eqref{eq:ctr} the oracle season values are
\$804.3 (baseline), \$747.2 (low mean), and \$686.1 (low dispersion).
The oracle bids on--off with switching curve $\swc{t} \approx t/2$ and
posts a harvest discount $d_H = 5.4$ at the baseline (a \$64.60 bundle,
with the vase at \$14.60) against clearance-deep discounts in the weak
markets, where bid-only revenue collapses to the passive level while the
joint policy earns 13 to 14 times as much, the complementarity of
\citet{Mohammadi2026} recomputed under the anchored curve.

Conditions run paired replicates under common random numbers (identical
customer uniforms per replicate and season), $\reps = 200$
unless Table~\ref{tab:design} says otherwise; the agent, \textsc{psrl},
\textsc{ce} and $\varepsilon$-greedy run $2{,}000$ replicates in the four
main settings. Regret is computed as Section~\ref{sec:learning} describes:
exactly for learners that play one plan per season, against the oracle
under common random numbers for the others. Differences between learners
are reported as paired means with 95\% confidence intervals.

\begin{table}[!t]
\centering
\caption{Calibration and experimental design.}
\label{tab:design}
\small
\begin{tabular}{@{}l>{\raggedright\arraybackslash}p{12.4cm}@{}}
\toprule
Season & $T = 20$, $I_0 = 15$, $s = 0$; $b_{\max} = d_{\max} = 15$, controls on $0.1$-grid \\
Prices, tastes & $p_P = 50$, $p_S = 20$; $X \sim \mathrm{Gamma}(40, 0.1)$,
$R_S \sim \mathrm{Gamma}(5, 4)$; $k = 0.05$ \\
Click curve & $\lambda_0 = 0.3$, $\lambda_\infty = 1$, $\ctrslope = 1$;
$\ctroffset$ unknown \\
Grid $\parmspace$ & $\delta_\mu \in \{50,\dots,170\} \times \delta_\sigma
\in \{2,\dots,42\} \times \ctroffset \in \{4,\dots,7\}$, $J = 125$ \\
Truths & baseline $(170,42;5.5)$; low mean $(50,6;5.5)$; low dispersion
$(60,2;5.5)$ \\
Replicates & $\reps = 200$, paired; $2{,}000$ for the four main learners in
the main settings, $100$ in robustness and scaling; seeds $(\text{seed},
\text{replicate}, \text{season})$ \\
Seasons & headline, weak markets $\nseasons = 500$; trap $1000$; waiting
$300$; robustness $300$ \\
Trap & truth $(170, 42; 4)$; prior mass $0.98$ ($0.95$, $0.90$) on an
added unresponsive curve, $\ctroffset = 20$ \\
\bottomrule
\end{tabular}
\end{table}

\subsection{Headline: regret against the clairvoyant}
\label{sec:headline}

\begin{figure}[!t]
\centering
\includegraphics[width=0.92\textwidth]{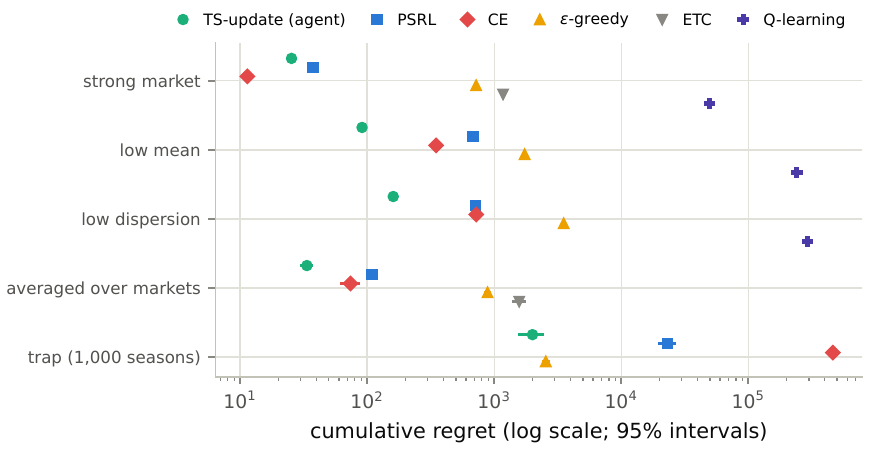}
\caption{Cumulative regret over 500 seasons (the trap: 1,000) with 95\%
intervals. The agent, $\varepsilon$-greedy and Q-learning are scored
against the oracle under common random numbers, the others exactly.
}
\label{fig:regret}
\end{figure}

Figure~\ref{fig:regret} sets the learners side by side in every test case
and Table~\ref{tab:headline} gives the numbers at the strong baseline truth
and averaged over the prior. The agent (\textsc{ts-update}) has the lowest
regret in four of the five settings. Averaged over the prior (Bayes mode,
the truth drawn from the prior) it gives up $33.6$ over 500 seasons, below
\textsc{ce} by $40.4$ (paired 95\% interval $[29.7, 51.1]$) and below
\textsc{psrl} by $75.8$ ($[67.7, 84.0]$). At the strong baseline truth it
gives up $25.4$, about 3\% of one season's oracle revenue of \$804.3, and
$22.1$ of that in the first ten seasons; drawing once per season gives up
$12.0$ more ($[9.9, 14.2]$). Certainty equivalence does better at this
truth, at $11.4$, below the agent by $14.0$ ($[12.0, 15.9]$). With a
uniform prior its mean model already bids, so its clicks keep correcting
its beliefs, and it never plays a deliberately sampled plan, so it pays
nothing for exploration. The trap of Proposition~\ref{prop:trap} stays
silent here.

Exploration without the model is expensive. Explore-then-commit pays
$1{,}175$, nearly all of it in its ten sweep seasons, and decaying
$\varepsilon$-greedy keeps paying for exploration long after it stops
learning ($722$). The model-free bandit, Thompson sampling over $256$
fixed bid--discount pairs with the season's revenue as its reward, pays
$63{,}190$ and still sits 6.5\% below the oracle after 500 seasons.
Tabular Q-learning needs no demand model at all and pays for it: it gives
up $49{,}456$ over 500 seasons at the baseline and still earns 9\% less
than the oracle at the end, and in the weak markets it ends at 45\% and
18\% of the oracle's revenue. Run for 20,000 seasons, its greedy policy
reaches 95\% of the oracle at the baseline after 6,500 seasons and stays
near 82\% in the low-mean market (Table~S.5).
Which half of the truth is costly to learn depends on the market.
At the saturated baseline it is the click side: knowing the willingness to
pay still leaves $31.0$, knowing the click curve only $6.5$. Averaged over
the prior it is the willingness-to-pay side ($65.3$ against $37.1$). These
half oracles draw once per season.

\begin{table}[!t]
\centering
\caption{Cumulative regret. The first four learners: $2{,}000$
replicates; the others: $200$. The agent, $\varepsilon$-greedy and
Q-learning are scored against the oracle under common random numbers.
Bayes mode draws the truth from the prior; the half oracles draw once per
season.
}
\label{tab:headline}
\small
\begin{tabular}{lrrr}
\toprule
 & \multicolumn{2}{c}{Baseline truth} & Bayes mode \\
\cmidrule(lr){2-3}\cmidrule(lr){4-4}
Policy & 10 seasons & 500 seasons (s.e.) & 500 seasons (s.e.) \\
\midrule
Agent (\textsc{ts-update}) & 22.1 & 25.4 (1.0) & 33.6 (1.6) \\
One draw per season (\textsc{psrl}) & 33.0 & 37.4 (0.6) & 109.4 (4.5) \\
Certainty equivalence & 9.5 & 11.4 (0.3) & 74.0 (5.6) \\
$\varepsilon$-greedy & 47.7 & 721.5 (3.5) & 887.8 (13.7) \\
Explore-then-commit & 1{,}169.6 & 1{,}175.4 (0.6) & 1{,}574.2 (80.8) \\
Thompson sampling, 256 fixed pairs & 2{,}007.1 & 63{,}190.1 (45.5) & 72{,}904.8 (2{,}151.3) \\
Tabular Q-learning, 36 pairs & 2{,}790.0 & 49{,}456.1 (398.7) & \\
Known click curve (learn WTP) & 6.5 & 6.5 (1.3) & 65.3 (13.5) \\
Known WTP (learn click curve) & 27.2 & 31.0 (1.2) & 37.1 (2.5) \\
Best static pair (clairvoyant) & & 4{,}674.1 & 7{,}442.6 \\
Passive & & 225{,}716.6 & \\
\bottomrule
\end{tabular}
\end{table}

\subsection{Recovering the switching curve}
\label{sec:recovery}

\begin{figure}[!t]
\centering
\includegraphics[width=0.85\textwidth]{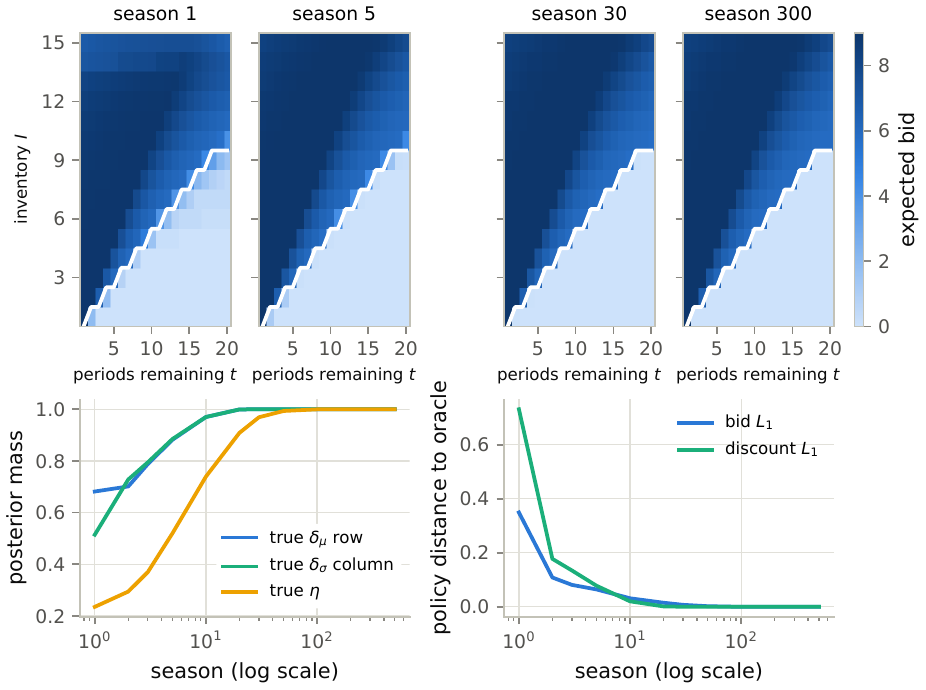}
\caption{Policy recovery at the low-mean truth under \textsc{psrl}. Top:
posterior-expected bid policy after $n$ seasons (white: oracle switching
curve). Bottom: posterior mass on the true parameters (left) and distance
to the oracle policies (right).
}
\label{fig:recovery}
\end{figure}

Figure~\ref{fig:recovery} shows \emph{what} is learned, at the harder
low-mean truth under \textsc{psrl}. The posterior-expected bid policy is
already recognizably on--off after one season and visually
indistinguishable from the oracle by season 30. The mean $L_1$ distance
between the expected and oracle bid policies falls from
$0.35$ (season 1) to $0.0075$ (season 30) to numerical zero by season 200.
Posterior mass on the true market-strength row reaches $0.88$ within
five seasons and $1.00$ by season 30, because season-start states
separate weak from strong almost perfectly; the click-curve location,
learnable only through bids, lags (mass $0.91$ at season 20, $1.00$ by
season 200). At the baseline truth the posterior \emph{never}
concentrates, yet regret vanishes, as Propositions~\ref{prop:harvest} and
\ref{prop:saturated} predict: the truth and the nine other saturated
market grid points prescribe the same season plan, and after $500$
seasons they hold $0.595$ of the posterior in ten \emph{equal} shares of
$0.0595$ (S1). At the low-mean truth the class holds $0.011$ of the
posterior after one season and $5 \times 10^{-22}$ after two
(Proposition~\ref{prop:saturated}(iv)).

\subsection{Weak markets and the trap}
\label{sec:weak}

\begin{figure}[!t]
\centering
\includegraphics[width=0.92\textwidth]{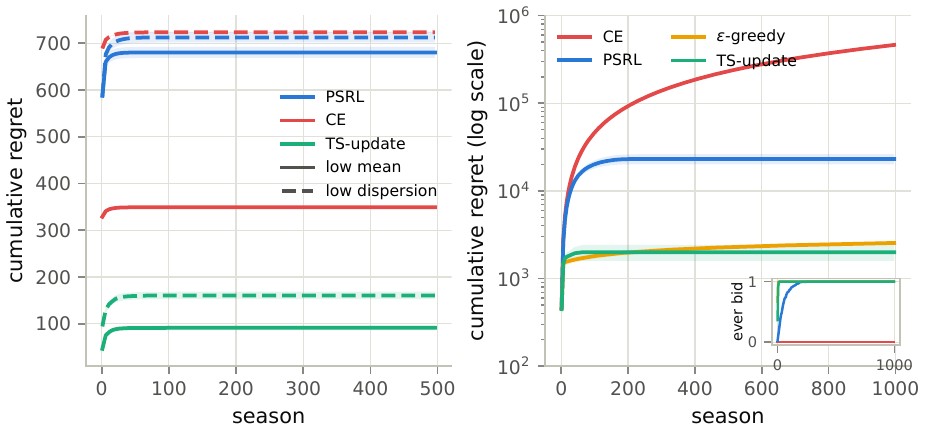}
\caption{Left: cumulative regret at the two weak-market truths. Right:
the trap, under prior mass $0.98$ on an unresponsive click curve (log
scale); inset: share of replicates that have ever bid.
}
\label{fig:weak}
\end{figure}

Weak markets are where learning costs money. At the low-mean truth, where
bid-only revenue collapses to the passive level ($52.9$ against the joint
oracle's $747.2$), the agent gives up $91.3$ over 500 seasons, and $160.7$
at the low-dispersion truth, about 12\% and 20\% of one season's oracle
revenue (Figure~\ref{fig:weak}, left). It earns $99.7\%$ and $99.5\%$ of
the oracle over its first fifty seasons, more than the best static pair
chosen with full knowledge of demand ($92.5\%$ and $93.4\%$; both in
Table~S.3).
One draw per season costs four to seven times as much ($680.5$ and
$712.5$). Certainty equivalence lies in between at the low mean
($349.2$) and above both at the low dispersion ($723.7$), and
$\varepsilon$-greedy pays $1{,}736$ and $3{,}528$. The gap between the two
sampling schedules comes from the first seasons. A draw from a strong
market posts harvest discounts that a weak market cannot convert, and the
first no-sales at those discounts already refute it. \textsc{ts-update}
acts on that evidence in the next period, \textsc{psrl} only in the next
season.

The right panel of Figure~\ref{fig:weak} shows the trap of
Proposition~\ref{prop:trap} at a strong market with cheap clicks
($\ctroffset = 4$, oracle value $825.9$), under a prior with mass $0.98$
on a click curve whose midpoint lies beyond the feasible bid range, so
that the mean-model bid is zero in every state.
Certainty equivalence never bids in any of $200$ replicates over $1{,}000$
seasons and loses $464.9$ per season, more than half of the oracle's
revenue, so its cumulative regret grows linearly. \textsc{psrl} draws
a responsive click curve with probability $0.02$ per season. It first
bids in season $50.9$ on average (predicted: $50$) and gives up $23{,}210$
in total. \textsc{ts-update} draws twenty times a season, first bids in
season $2.9$ on average (predicted: $3.0$) and gives up $2{,}004$. At live
mass $0.05$ and $0.10$ the mean first-bid seasons are $21.1$ and $10.0$
for \textsc{psrl} and $1.5$ and $1.1$ for \textsc{ts-update}, against
predicted $20$, $10$, $1.6$ and $1.1$ (Figure~S.1). Randomized
exploration also escapes ($\varepsilon$-greedy: $2{,}552$) but keeps
paying its exploration tax where nothing is left to learn ($722$ at the
baseline). A point-estimate system that recommends ``do not advertise''
produces data that can never contradict it.

\subsection{How the results compare with existing findings}
\label{sec:discussion}

Certainty equivalence does well wherever its own decisions keep producing
informative data.
\citet{denBoerZwart2015} show that with finite inventories and repeated
selling seasons a slightly modified certainty-equivalent pricing policy
learns, because optimal prices vary with the remaining stock. Here the
bid switches on and off with stock and time in the same way, so clicks
keep informing the click-curve estimate, which is why \textsc{ce} wins at
the strong baseline. That endogenous learning stops once the plan stops
bidding, because a zero bid makes clicks uninformative. The trap is the
incomplete learning of \citet{Harrison2012} and \citet{Keskin2014} moved
from the price to the bid, and the remedy changes with it: those papers
force deviations from the myopic price, while here a single positive bid
informs, so one responsive draw suffices. The Thompson-sampling algorithms of
\citet{Ferreira2018} also sample every period and re-solve a
linear program each period, in one variant with inventory constraints
that follow the remaining stock.
With exactly solved plans the stock and time are already handled, and the
sampling schedule is the design choice left. It matters most where the
first observations contradict the draw, in weak markets and in the trap.
\citet{Gurkan2026} give sufficient conditions for exponentially
fast learning when price, advertising and inventory are set together, and
find that departing from them to avoid experimentation costs profit. A
zero bid is the extreme case: it avoids experimentation on the
advertising side altogether, and learning about clicks stops. Model-free
agents, such as the
reinforcement-learning pricers of \citet{Zhou2022} and \citet{Qiao2024},
learn a policy from many simulated episodes. The two model-free learners
here show what that costs without the season structure: the bandit over
fixed pairs gives up $63{,}190$ and tabular Q-learning $49{,}456$ over 500
seasons, and Q-learning needs thousands of seasons to come within 5\% of
the oracle. A model-based agent avoids that cost by learning three
parameters rather than a value function. Section~\ref{sec:strategic}
adds two comparisons on the customer side.

\subsection{Prior misspecification and robustness}
\label{sec:robust}

In every robustness variant the agent either beats once-per-season
sampling or matches it within sampling error (Table~S.2). An off-grid truth
leaves a regret plateau that tracks the mesh: a roughly
sevenfold finer grid cuts it from $0.42$ to $0.10$ per season
(Table~S.2).
The picture is unchanged across $(T, I_0)$ and positive salvage, but not
when the display sensitivity $k$ is learned as well: at a fast-decay truth
off a coarse $81$-point grid posterior sampling settles at $90.2\%$ of
the oracle.
When the market weakens every 40 seasons, every learner that keeps all
its data pays once the market turns weak, and keeping only the last 40
seasons of data cuts the regret by a third (S2).
Regret is flat in the grid size ($36.6$ at $J = 27$, $31.9$ at $J = 405$;
Figure~S.1), far from the $\sqrt{J \log J}$ of
Proposition~\ref{thm:regret}, because every observation updates all
candidates at once.

\section{Customers who wait for a better discount}
\label{sec:strategic}

\subsection{Waiting on the discount margin}
\label{sec:waitmodel}

Component prices are fixed within a season, so the only price a customer
can hope to improve by waiting is the bundle's. Following the three-way
choice model of \citet{Mohammadi2026b}, itself adopted from
\citet{Nabi2018}, a clicking customer whose
surplus-maximizing option is the bundle, with surplus $u = R_P + R_S -
(p_P + p_S - d_t) \ge 0$, compares buying now, worth $u$, with waiting one
period for a deeper discount, worth $\pat\,(u + \dest_t - d_t)$ at
patience $\pat \in [0, 1)$ and a forecast $\dest_t$ of the posted
discount. The customer therefore waits exactly when
\begin{equation}
\label{eq:wait}
u \;<\; \wthr_t \;=\; \frac{\pat\,\bigl(\dest_t - d_t\bigr)^{+}}{1 - \pat},
\end{equation}
buys otherwise, and, having waited, returns next period with probability
$\retfrac$, purchasing its surplus-best option at the then-posted discount
and displayed inventory. As in \citet{Mohammadi2026b}, the forecast
weights recent prices most. Here it exponentially smooths the posted
discount within the season, $\dest_{t-1} = (1 - \fweight)\,\dest_t +
\fweight\, d_t$, starting afresh at each season's first posted discount.
Customers thus hold adaptive expectations: they respond to the posted price path
and never see the seller's algorithm, and they wait when the discount
falls below its recent level. This is a scope choice. A discount that
deepens through the season, the classic markdown, never makes them wait,
whereas customers with rational expectations would wait for it
\citep{Aviv2008,Su2008}. Setting $\pat = 0$ recovers the myopic
environment of Section~\ref{sec:model} exactly.

Waiting is priced entirely by the discount path:

\begin{proposition}[Immunity of constant discount paths]
\label{prop:immunity}
Under the waiting rule \eqref{eq:wait}: (i) $\wthr_t$ is nondecreasing in
$\pat$ and in the forecast gap $(\dest_t - d_t)^{+}$ and zero whenever
$d_t \ge \dest_t$. (ii) If the posted discount is constant within the
season, $d_t \equiv d$, then $\dest_t = d$ and $\wthr_t = 0$ in every
period: no customer waits, revenue and transitions coincide with the
myopic environment's for every $\pat \in [0, 1)$ and every bid path, and
hence the season value of any policy posting season-constant discounts is
invariant to $\pat$. (iii) In a saturated market the full-information
plan is such a policy, since by Proposition~\ref{prop:saturated}(ii) it
posts the harvest discount $d_H$ in every state: every saturated market's
oracle is \emph{immune to waiting}, and any waiting observed there
under a learner is caused by whoever varies the discount, that is, by
learning itself.
\end{proposition}

Parts (i) and (ii) follow directly from \eqref{eq:wait}, and (ii) carries
to the discount the observation of \citet{Mohammadi2026b} that a price
held constant leaves customers nothing to wait for. Part (iii) ties
immunity to the saturated class. Off that class the oracle's own discount
moves with the state, taking $37$ distinct levels at the low-mean truth
and $43$ at the low-dispersion truth, so there even the clairvoyant seller
trains waiting.

The seller's likelihood stays myopic: a customer who waits is recorded as
a click without a sale, indistinguishable from a genuine $\emptyset$, and
returning purchases arrive outside the click-attributed data stream.
Waiting therefore also \emph{contaminates the seller's beliefs}, with a
sharp threshold:

\begin{lemma}[A dichotomy for belief contamination]
\label{lem:contamination}
Fix two WTP candidates, the truth $\parmw^{*}$ and a rival $\parmw'$,
both with all four outcome probabilities positive at every visited state,
and let each
clicking customer whose surplus-best option is the bundle wait with
probability $w$. (i) At any state $(I, d)$ the expected per-click
increment of the log-likelihood ratio of $\parmw'$ against $\parmw^{*}$
is $-\KL{\alpha(\parmw^{*})}{\alpha(\parmw')} + w\,S_{I,d}$, with slope
\[
S_{I,d} \;=\; \aB(\parmw^{*}) \log
\frac{\aN(\parmw')\, \aB(\parmw^{*})}{\aN(\parmw^{*})\, \aB(\parmw')};
\]
it is linear in $w$, negative at
$w = 0$, and, whenever $S_{I,d} > 0$ (for which $\aB(\parmw') <
\aB(\parmw^{*})$ and $\aN(\parmw') > \aN(\parmw^{*})$ at that state
suffice), positive exactly above $w^{*}_{I,d} =
\KL{\alpha(\parmw^{*})}{\alpha(\parmw')} / S_{I,d}$. (ii) Aggregate over
a frozen season plan's visited states, weighting by expected clicks with
the visit measure held at its myopic benchmark and seasons i.i.d., and
suppose the aggregated slope $\overline{S}$ is positive, a computable
condition on known quantities. Then the posterior odds of $\parmw'$
against $\parmw^{*}$ converge to $0$ almost surely when the waiting rate
is below $\bar{w}^{*} = \overline{\mathrm{KL}} / \overline{S}$ and
diverge almost surely when it is above.
\end{lemma}

Waited bundles become recorded no-sales, so beliefs drift toward weaker
markets once the waiting rate crosses the threshold. This is the
learning counterpart of the spiral-down effect of \citet{Cooper2006}, in
which a revenue manager who ignores customers' substitution reads it as
weaker demand. Under the frozen clairvoyant plan at the weak truth only
the weakest rival $(50, 2)$ has a positive aggregated slope, with
$\bar{w}^{*} = 0.275$.
Forcing exogenous waiting on that plan keeps posterior mass on the truth
at $1.00$ up to $w = 0.32$ and drives it to $0.00$ from $w = 0.45$, so for
a frozen plan the computed threshold is conservative (S1, Figure~S.2).
For a learner, whose plans vary, it is a guide only: at $\pat = 0.9$ the
season-level learner's realized waiting rate is about $0.27$, just below
the threshold, and its mass on the truth has already fallen to $0.33$.

\subsection{Results}
\label{sec:strategicresults}

\begin{figure}[!t]
\centering
\includegraphics[width=0.92\textwidth]{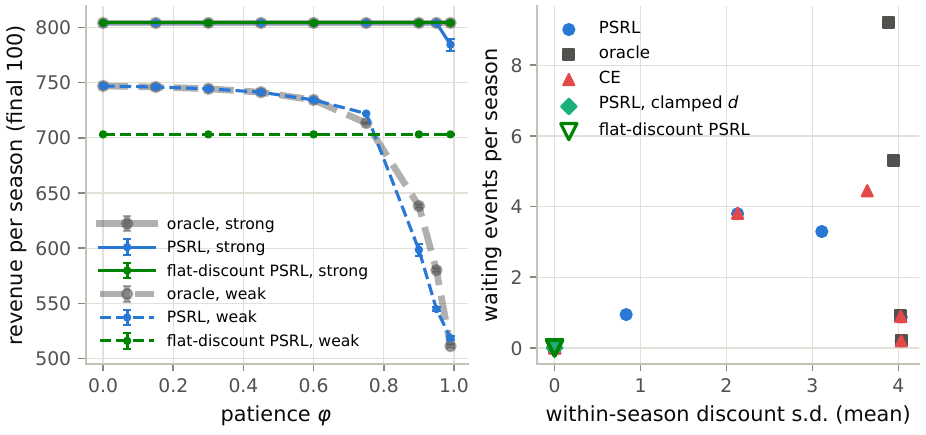}
\caption{Customers who wait. Left: revenue per season (final 100
seasons) against patience $\pat$. Right: waiting incidence against the
within-season standard deviation of the discount.
}
\label{fig:patience}
\end{figure}

Figure~\ref{fig:patience} reports the waiting experiments, and
Table~S.4 gives the weak-truth results by patience level. The protocol of
Section~\ref{sec:setup} applies, with $\retfrac = 0.5$ and $\fweight =
0.3$. Alongside the learners of Section~\ref{sec:results} we run the
\emph{flat-discount learner}, the variant of \textsc{psrl} that
Proposition~\ref{prop:immunity}(ii) suggests: draw $\parm$ from the
posterior once per season, then play the best season plan whose discount
is one constant level (the level is optimized at the draw and the bid
stays state-dependent).

\paragraph{Exposure to waiting}
At the strong-market truth the oracle's simulated revenue is $804.1$ per
season (exact value $804.3$) at \emph{every} patience level, because the
flat harvest discount gives customers nothing to wait for. Both sampling
learners match it for $\pat \le 0.9$, since their discount settles within
a few seasons. At $\pat = 0.99$ early exploration trains waiting ($0.94$
events per season for \textsc{psrl}, $1.10$ for \textsc{ts-update}), and
both lose $2.5\%$ of revenue on average while the oracle loses nothing.
The average mixes two outcomes. In 150 (\textsc{psrl}) and 137
(\textsc{ts-update}) of the 200 runs no customer ever waits and nothing is
lost; in the rest an early wait refutes the saturated class, and revenue
settles at $90.2\%$ and $92.2\%$ of the oracle's (S1).
weak-market truth the oracle itself discounts less as stock clears, so
all are exposed. At $\pat = 0.9$ the oracle reaches $85.4\%$ of the
myopic-clairvoyant value, \textsc{ce} $83.7\%$, \textsc{psrl} $80.1\%$
and \textsc{ts-update} $79.5\%$, and at $\pat = 0.99$ the three learners
all reach $69.4\%$. Drawing every period buys nothing here: once the
posterior settles, both schedules post the same plans.

\paragraph{What trains waiting}
Waiting needs a discount that varies within the season. Where it does
not vary, nobody waits; where it does, patience drives how many do
(Figure~\ref{fig:patience}, right). Clamping the learner's discount (at the cap $d_{\max} = 15$ at the
weak truth and at $d_H = 5.4$ at the strong truth) while the bid
keeps learning eliminates waiting at both truths and, at the weak market
with $\pat = 0.9$, recovers revenue to $93.6\%$ of the clairvoyant, above
both the unclamped learners and the oracle. Waiting also contaminates
inference. At the weak truth, posterior mass on the true $\parmw$ after
$300$ seasons falls from $1.00$ ($\pat \le 0.3$) to $0.98$, $0.33$ and
$0.00$ as $\pat$ rises to $0.6$, $0.9$ and $0.99$ under \textsc{psrl}, and
to $0.29$ and $0.00$ at the two highest levels under \textsc{ts-update}.
At the saturated truth a waited bundle is the no-purchase
observation that every saturated parameter rules out
(Proposition~\ref{prop:saturated}(iv)), so a single wait refutes the whole
class (S1).

\paragraph{Committing the discount}
The flat-discount learner turns Proposition~\ref{prop:immunity}(ii) into a
design rule. Its waiting count is zero at every patience level by
construction, and its beliefs stay clean: posterior mass on the true
$\parmw$ at the weak truth is $1.00$ for every $\pat$. The price is the
flatness itself. At the weak truth the best flat plan earns $94.1\%$ of
the clairvoyant value, so under myopic customers commitment gives up
$5.9$ points that the unrestricted learners keep. At $\pat = 0.9$ it earns
more than the unrestricted learners, the learner with a clamped discount
and the myopic-optimal oracle, and at $\pat = 0.99$ it leads the
unrestricted learners by $24.7$ points ($94.1\%$ against $69.4\%$). At the strong truth commitment is free: the optimal plan is
already flat (Proposition~\ref{prop:saturated}(ii)), and the flat-discount
learner earns the oracle's simulated $804.1$ at every patience level.
Committing the discount need not slow the bid. An agent that keeps drawing
the bid every period but holds the discount drawn at the start of the
season makes nobody wait either, earns $94.1\%$ of the clairvoyant value
in the weak market at every patience level and, in the trap, first bids
after $2.9$ seasons on average, as the agent does. Its price is a slower
start on the discount, about $800$ in the weak markets on top of the cost
of flatness, most of it in the first ten seasons (S2).
seller can watch for waiting directly: returning customers buy outside
the click-attributed data, so bundle purchases without a click that rise
after the discount falls are its signature. Committing the discount
mirrors \citet{AvivWei2019}, who find that committing to a price
increasingly dominates responsive pricing as strategic customers become
more common, and it keeps learning across seasons while committing within
each.

\section{Rules for deployment}
\label{sec:managerial}

Table~\ref{tab:guidelines} turns the results into six rules for a seller
who launches a perishable product into an unknown market, each with a
signal the seller can watch. The numbers come from the simulated florist
of Section~\ref{sec:setup}.

\begin{table}[!t]
\centering
\caption{Rules for deploying the agent.}
\label{tab:guidelines}
\footnotesize
\setlength{\tabcolsep}{3pt}
\newcolumntype{L}[1]{>{\raggedright\arraybackslash}p{#1}}
\begin{tabular}{@{}L{2.9cm}L{3.3cm}L{4.3cm}L{4.9cm}@{}}
\toprule
Rule & Signal to watch & Action & Evidence here \\
\midrule
Draw every period & always & draw a market from the posterior in each
period and post its plan's bid and discount & regret 91.3 and 160.7 in the
weak markets against 680.5 and 712.5 with one draw per season; 33.6
against 109.4 averaged over markets (Sections~\ref{sec:headline}
and~\ref{sec:weak}) \\
\addlinespace
Keep a prior floor on bidding & before launch & put prior mass
$m \ge 1 - \delta^{1/(TK)}$ on click curves whose plans bid & $m \ge 0.049$
bids within three seasons with probability $0.95$
(Proposition~\ref{prop:trap}); without bidding, \$464.9 is lost every
season \\
\addlinespace
Distrust a plan to stop advertising & the posterior-mean plan bids zero
everywhere & keep sampling until a bid has moved the click-curve
posterior & a zero bid reveals nothing about paid clicks
(Lemma~\ref{lem:censoring}) \\
\addlinespace
Freeze the discount where scarcity carries the price & the estimated
share of full-stock shoppers who would not buy the product at its list
price is below a small $\varepsilon$ & post the harvest discount ($\$5.40$
here) and learn only the bid & the loss is at most of order
$\varepsilon$ (Proposition~\ref{prop:harvest}); tuning buys nothing there \\
\addlinespace
Commit the discount when shoppers wait & bundle purchases without a click
rise after the discount falls & hold the discount drawn at the start of
the season and keep drawing the bid every period & $94.1\%$ of clairvoyant
revenue at every patience level, against $80.1\%$ and $69.4\%$ at
patience $0.9$ and $0.99$; the bid still leaves the trap after $2.9$
seasons; costs $5.9$ points when nobody waits, nothing in the strong
market (Section~\ref{sec:strategicresults}) \\
\addlinespace
Refine the grid when learning stalls & the posterior settles between
neighbouring candidates & add candidates near the posterior mode and
re-solve offline & a sevenfold finer grid cuts the late regret from $0.42$
to $0.10$ per season (Table~S.2) \\
\bottomrule
\end{tabular}
\end{table}

Two of the rules have a price. The prior floor makes the agent bid in
seasons where a pessimistic seller would not, which costs money if clicks
really are unresponsive. Commitment gives up the discount's response to
stock: at patience $0.3$ and $0.6$ the unrestricted learner still earns
$99.6\%$ and $98.3\%$ of the clairvoyant value, so committing would cost
$5.5$ and $4.2$ points there, and it pays only once patience is high.

\section{Conclusions}
\label{sec:conclusion}

We designed and tested a learning agent for an online retailer that must
learn how clicks respond to its sponsored-search bid and what shoppers will
pay for a perishable product as displayed stock runs down. Sampling a
candidate market every period and following its exactly solved plan learns
fastest where learning is expensive, escapes a planner's decision to stop
advertising within a few seasons, and leaves every action readable. Where
displayed scarcity keeps willingness to pay above the posted price, the
discount needs no learning, and where shoppers wait for a better discount,
one discount per season keeps their waits from biasing the agent.

\subsection*{Limitations and future research}

The evidence is simulated from an illustrative calibration, so a field
test on a retailer's click and sales logs is the natural next step. Three
modeling choices deserve that test. Every click is billed at the bid,
organic ones included, which overstates the cost of bidding. Customers
who wait hold adaptive expectations, so a classic markdown that deepens
through the season does not make them wait. Customers who anticipate the
seller's plan would, and a learner facing them would need a model of
their timing. The scarcity response is assumed: scarcity cues raise
booking intention on hospitality platforms \citep{Teubner2020}, which
supports the assumption, but its exponential form and the display
sensitivity $k$ are taken as known. When $k$ is learned as well, at a
fast-decay truth off a coarse grid, posterior sampling settles at
$90.2\%$ of the oracle (Section~\ref{sec:robust}).
The parameter grid is finite. Refining it is cheap and the
misspecification plateau tracks the mesh, but fully continuous posteriors
would need approximate season solves. Component prices are fixed within
the season, one product pair is bundled, and the stock is displayed in
every period, although whether and when to disclose it is itself a
decision \citep{Liu2026}. Competition enters only through a stationary
click curve for a single keyword. A learner over several keywords would
answer the call of \citet{LiYang2022} for keyword targeting that adjusts
to market responses. Our model-free benchmarks are a bandit over fixed
pairs and tabular Q-learning; deep reinforcement learning, which needs
simulated training at a scale the agent avoids, is left for problems with
richer state. On the theory side, a regret bound for sampling within
the season and instance-dependent regret under censoring remain
unresolved.

\appendix
\section{Proofs}
\label{app:proofs}

This appendix proves the claims of Section~\ref{sec:oracle} and the
results of Sections~\ref{sec:theory} and~\ref{sec:strategic}, in the
order in which they appear.

\paragraph{Separation and monotonicity of the season program (Section~\ref{sec:oracle})}
(i) In \eqref{eq:bellman} the bid enters only through $\ctr{b}$ and the
per-click cost, and $\inner_I$ depends on $(I, d)$ alone, so
$\max_{b,d}\ctr{b}[\inner_I(d; \Delta, \parm) - b] = \max_b
\ctr{b}[\inner_I^*(\Delta, \parm) - b]$ with $\inner_I^* = \max_d \inner_I$.
(ii) $\inner_I^*$ is a maximum of functions affine and nonincreasing in
$\Delta$ (the coefficient of $-\Delta$ is $\aP + \aB \ge 0$), hence
nonincreasing in $\Delta$. As $\ctr{\cdot}$ is nondecreasing,
$\ctr{b}(v - b)$ has increasing differences in $(b, v)$, so by Topkis's
theorem \citep{Topkis1998} the largest and smallest maximizers $b^*$ are
nondecreasing in $v = \inner_I^*(\Delta)$, hence nonincreasing in $\Delta$.
A deeper discount raises only the bundle's surplus, so the no-purchase and
secondary-only events shrink: $\aP + \aB = 1 - \aS - \aN$ is
nondecreasing in $d$, $\inner_I$ has decreasing differences in $(d,
\Delta)$, and $d^*$ is nonincreasing in $\Delta$.\qed

\paragraph{Proof of Proposition~\ref{prop:harvest}}
Write $F \equiv F_P(\cdot \mid I; \parmw)$ and let $\varepsilon =
F(p_P)$: the probability that the primary reservation price fails to
clear the undiscounted primary price. On the event $R_P > p_P$, the
primary-alone surplus is positive, the bundle beats the secondary alone
(their surplus difference is $R_P - (p_P - d) > 0$), and the bundle beats
the primary alone exactly when $R_S > p_S - d$; so the choice is the
bundle with probability $\bar{F}_S(p_S - d)$ and the primary alone
otherwise, regardless of the fine location of $R_P$. Hence
$|\aS| \le \varepsilon$, $|\aN| \le \varepsilon$,
$|\aP - F_S(p_S - d)| \le \varepsilon$, and
$|\aB - \bar{F}_S(p_S - d)| \le \varepsilon$, uniformly in $d$.
Substituting into $\inner_I(d; \Delta, \parm) = (\aP + \aB)(p_P - \Delta)
+ (\aS + \aB)\,p_S - \aB\,d$,
\[
\bigl|\inner_I(d; \Delta, \parm) - \bigl[(p_P - \Delta) +
\bar{F}_S(p_S - d)(p_S - d)\bigr]\bigr|
\;\le\; (|p_P - \Delta| + 2p_S + d_{\max})\,2\varepsilon,
\]
using $\aS + \aB = \aB + O(\varepsilon)$ and $\bar F_S(p_S-d)\,p_S -
\bar F_S(p_S-d)\,d = \harv(d)$ after collecting terms. The bound is
uniform on compact $\Delta$-sets and vanishes as $\varepsilon \to 0$. The
limit objective is $(p_P - \Delta) + \harv(d)$, whose $d$-dependence is
all in $\harv(d) = (p_S - d)\bar{F}_S(p_S - d)$, so every
selection of maximizers converges to $d_H = \arg\max_d \harv(d)$
(unique at the calibration), which involves only the known $F_S$.\qed

\paragraph{Proof of Lemma~\ref{lem:censoring}}
(i) By the anchored curve \eqref{eq:ctr}, $\ctrtheta{0}{\parmc} =
\lambda_0$ for every $\parmc$, so with $b_t \equiv 0$ every click factor
in \eqref{eq:factorization} equals $\lambda_0^{c}(1 - \lambda_0)^{1-c}$,
independent of $\parmc$, so under a product prior the $\parmc$-posterior
never moves.

(ii) Fix $(I, d)$ and couple the two models through a common $R_S$ draw.
Condition on $R_S = r$ and trace the choice as a function of $R_P$. Two
cases exhaust the possibilities.
\emph{Case $r < p_S - d$.} The bundle's surplus is below the primary's
($s_B - s_P = r - (p_S - d) < 0$) and the secondary's is negative
($s_S = r - p_S < -d \le 0$), so the choice is $P$ if $R_P > p_P$ and
$\emptyset$ otherwise: one decisive threshold, $p_P \in W(d)$.
\emph{Case $r \ge p_S - d$.} The bundle weakly dominates the primary,
so the choice is among $B$, $S$, $\emptyset$: the bundle beats the
secondary alone exactly when $s_B > s_S$, i.e.\ $R_P > p_P - d$, and
beats no-purchase exactly when $R_P > p_P + p_S - d - r$; when the
bundle loses both comparisons the choice is $S$ if $r > p_S$ and
$\emptyset$ otherwise, with no further dependence on $R_P$. The two
decisive thresholds are $p_P - d$ and $p_P + p_S - d - r$, and the
latter matters only while it exceeds $p_P - d$, i.e.\ for $r \le p_S$,
and is at most $p_P$ because $r \ge p_S - d$: both lie in $W(d) =
[\,p_P - d,\; p_P\,]$.
In both cases the conditional choice law is a function of at most two
values of $F_P(\cdot \mid I)$ evaluated inside $W(d)$, so its total
variation across $\parmw, \parmw'$ is at most $2 \sup_{x \in W(d)}
|F_P(x \mid I; \parmw) - F_P(x \mid I; \parmw')|$. Integrating over the
common $r$ preserves the bound and gives the display in the lemma.

For the two inventory extremes, note $0 \le F_P(x \mid I; \cdot) \le
F_P(p_P \mid I; \cdot)$ on $W(d)$: if both distributions place at
most $\varepsilon$ of their mass below $p_P$, both distribution
functions are at most $\varepsilon$ throughout the window and the sup
difference is at most $\varepsilon$ (deep scarcity: every customer
clears every threshold and the choice reduces to the
secondary-membership comparison of Proposition~\ref{prop:harvest}).
Symmetrically, if both place at most $\varepsilon$ above $p_P - d$, both
distribution functions are at least $1 - \varepsilon$ on the window
(dead market: the primary never sells and the choice degenerates to
the secondary-alone comparison).\qed

\paragraph{Proof of Proposition~\ref{prop:saturated}}
Saturation gives $F_P(x \mid I; \parmw) = 0$ for every $x \le p_P$ and
every reachable $I$, hence also $f_P(\cdot \mid I) = 0$ there.
(i) Integrating the case analysis in the proof of
Lemma~\ref{lem:censoring} over $R_S$ gives $\aS = F_P(p_P - d
\mid I)\,\bar{F}_S(p_S) = 0$ because $d \ge 0$; $\aN = F_S(p_S) F_P(p_P -
d \mid I) + \int_{p_P - d}^{p_P} F_S(p_P + p_S - d - x) f_P(x \mid I)\,dx
= 0$ because the integration range lies in $(-\infty, p_P]$; $\aP = \bigl(1
- F_P(p_P \mid I)\bigr) F_S(p_S - d) = F_S(p_S - d)$; and $\aB = 1 - \aP$
by adding up. None of the four involves $\parmw$ or $I$.
(ii) Substituting $\aS = \aN = 0$, $\aP + \aB = 1$ into \eqref{eq:bellman}
leaves the inner objective $\ctr{b}\bigl[(p_P - \Delta) + \aB (p_S - d) -
b\bigr]$, and $\aB(p_S - d) = \harv(d)$ gives the stated recursion. Its
data are $p_P$, $F_S$ and the click curve only. With $\Vfun{I}{0} = I s$
and $\Vfun{0}{t} = 0$, also free of $\parmw$, the value function, both
maximizers (under the fixed tie-breaking rule) and therefore the whole
season policy are common to all saturated parameters, and the discount
maximizes $\harv$ alone, so it equals $d_H$ in every state.
(iii) By (i) the choice factor of the likelihood \eqref{eq:factorization} is
identical across $\parmspace_{\mathrm{sat}}$ for every observation, and
under a product prior the $\parmw$-posterior is the prior times the choice
factor, so the posterior odds between two saturated points never move from
their prior value. Renormalizing gives the claim. If $\parmsamp{n,w}$ and
$\parmtrue_w$ are both saturated then, by (ii), $\pi_n = \pi_{(\parmsamp{n,c},
\parmtrue_w)}$ and both the played and the optimal season value under
$\parmtrue$ are unchanged if $\parmsamp{n,w}$ is replaced by
$\parmtrue_w$: the season's regret is the click-curve sample's alone.
(iv) A post-click observation of $S$ or $\emptyset$ carries likelihood
factor $\aS = 0$ or $\aN = 0$ under every saturated parameter, so one such
observation sets the mass of the whole class to zero. Under a saturated
truth neither outcome ever occurs, while every non-saturated parameter
gives the observed outcomes positive probability, so it is down-weighted
only at its Kullback--Leibler rate.\qed

\paragraph{Proof of Proposition~\ref{prop:trap}}
(i) Suppose the posterior-mean model at the prior has optimal bid $0$ in
every state. Then \textsc{ce} posts $b_t \equiv 0$ in season 1. By
Lemma~\ref{lem:censoring}(i) the $\parmc$-posterior after season 1 equals
the prior. The $\parmw$-posterior may move (organic clicks still arrive
at rate $\lambda_0$), but by assumption the mean-model bid is zero in
every state \emph{for every} $\parmw$-posterior reachable under the
frozen $\parmc$-prior (the checkable premise). At our calibration it
holds because the pessimistic click curve makes paid traffic unprofitable
at every demand model on the grid. By induction \textsc{ce} never bids,
its click-curve beliefs never move, and it plays a no-bidding policy
every season. If the optimal no-bidding value under $\parmtrue$ falls
short of $\Vstar(\parmtrue)$ by $g > 0$, its regret is at least $g$ per
season. (ii) While no bid has been posted, the click posterior equals the
prior (Lemma~\ref{lem:censoring}(i)), and since the posterior is a
product, the click part of every draw has the prior law whatever its
willingness-to-pay part. By the premise a \textsc{psrl} draw bids during
the season exactly when its click curve is one whose plans bid, which has
probability $m$, and draws in different seasons use independent
randomness. The index of the first bidding season is therefore geometric
with success probability $m$. (iii) The same argument applies period by
period. \textsc{ts-update} draws afresh in every period, each draw's click
curve has the prior law until the first bid, and by the premise the draw
bids at the current state exactly when its click curve is one whose plans
bid. The first bidding period is therefore geometric with success
probability $m$. If a season without a bid lasts $T$ periods, it passes
without a bid with probability $(1 - m)^T$, independently across seasons,
so the first bidding season is geometric with success probability
$1 - (1 - m)^T = Tm + O\bigl((Tm)^2\bigr)$. After the first bid, click
probabilities at the posted bid differ across the click curves in the
prior's support, so the click-curve posterior moves.\qed

\paragraph{Proof of Proposition~\ref{thm:regret}}
Treat each season as one round of the Thompson-sampling problem of
\citet{Russo2016}. The action in season $n$ is the season plan $\pi_n \in
\Pi$, the observation $Y_n$ is the season's record of posted controls,
clicks and choices, and the reward is the season's revenue, a function of
$Y_n$ that lies in an interval of length $\bar{V}$. The set $\Pi$ is
finite, and each plan $\pi_\parm$ is a function of $\parm$ under the fixed
tie-breaking rule. Given $\parmtrue$ and the plan, the records of
different seasons are independent and identically distributed, because
every season restarts from $I_0$ under the same $\parmtrue$
(Section~\ref{sec:learning}). \textsc{psrl} draws $\parmsamp{n}$ from the
posterior $\post{n-1}$ and plays $\pi_n = \pi_{\parmsamp{n}}$. Since
$\parmsamp{n}$ and $\parmtrue$ have the same law given the history
$\hist{n-1}$,
\begin{equation}
\label{eq:bmatch}
\Prob\bigl(\pi_n = \pi \mid \hist{n-1}\bigr) \;=\; \Prob\bigl(\pi^{*} = \pi
\mid \hist{n-1}\bigr) \qquad \text{for every } \pi \in \Pi,
\end{equation}
so the algorithm is Thompson sampling over $\Pi$ in their sense (the
plans already played add no likelihood factor that involves $\parm$;
Section~\ref{sec:posterior}), and the regret \eqref{eq:regret} is their
Bayesian regret. Write $\E_n$ and $I_n$ for expectation and mutual
information given $\hist{n-1}$, and let
\[
\Gamma_n \;=\; \frac{\bigl(\E_n\bigl[V_{\parmtrue}(\pi^{*}) -
V_{\parmtrue}(\pi_n)\bigr]\bigr)^2}{I_n\bigl(\pi^{*};\, (\pi_n, Y_n)\bigr)}
\]
be the information ratio of season $n$. \citet{Russo2016} show that a
policy satisfying \eqref{eq:bmatch} has $\regret{\nseasons} \le
\sqrt{\bar\Gamma\, H(\pi^{*})\, \nseasons}$ whenever $\Gamma_n \le
\bar\Gamma$ in every season (their Proposition 1), and that $\Gamma_n \le
\lvert \Pi \rvert / 2$ when the actions are finitely many and the rewards
lie in an interval of unit length (their Proposition 3). Dividing the
revenue by $\bar{V}$ divides the numerator of $\Gamma_n$ by $\bar{V}^2$
and leaves the denominator unchanged, so $\Gamma_n \le \bar{V}^2 \lvert
\Pi \rvert / 2$, which gives the first bound. The second follows from
$H(\pi^{*}) \le \log \lvert \Pi \rvert$ and $\lvert \Pi \rvert \le
J$.\qed

\paragraph{Proof of Proposition~\ref{prop:immunity}}
(i) With surplus $u \ge 0$ and forecast gap $\delta = (\dest_t - d_t)^+$,
waiting is preferred exactly when $\pat(u + \delta) > u$, i.e.\ $u <
\pat\delta/(1 - \pat) = \wthr_t$; $\wthr_t$ is nondecreasing in $\pat$
and in $\delta$ and zero at $\delta = 0$. (ii) If $d_t \equiv d$ for all
$t$, the forecast recursion $\dest_{t-1} = (1 - \fweight)\dest_t +
\fweight d_t$, initialized at the first posted discount, gives $\dest_t
= d$ for all $t$, so $\delta \equiv 0$ and $\wthr_t \equiv 0$: no
customer waits, no returning pool forms, and revenue and transitions
coincide with the myopic environment's for every bid path, so the season
value of any season-constant policy is invariant to $\pat$. (iii) By
Proposition~\ref{prop:saturated}(ii) the full-information plan at a
saturated parameter posts $d_H$ in every state, a season-constant
discount, so (ii) applies to it.\qed

\paragraph{Proof of Lemma~\ref{lem:contamination}}
Fix the state $(I, d)$ and write $\alpha^{*} = \alpha(\cdot \mid I, d;
\parmw^{*})$, $\alpha' = \alpha(\cdot \mid I, d; \parmw')$ over the four
outcomes $\{P, S, \emptyset, B\}$. With waiting rate $w$, a clicked
customer produces the outcome law
\[
Q_w \;=\; \alpha^{*} + w\,\alpha^{*}_B\,(e_{\emptyset} - e_{B}),
\]
the truth's law with a $w$-fraction of bundle outcomes recorded as
no-sales. The myopic likelihood accumulates $\log \alpha_y$ at the
observed outcome $y$, so the expected per-click increment of
$\log(L'/L^{*})$ is
\[
D_{I,d}(w) \;=\; \sum_y Q_w(y) \log \frac{\alpha'_y}{\alpha^{*}_y}
\;=\; -\KL{\alpha^{*}}{\alpha'} \;+\; w\,\alpha^{*}_B \log
\frac{\alpha'_{\emptyset}\,\alpha^{*}_B}{\alpha^{*}_{\emptyset}\,
\alpha'_B},
\]
by direct substitution: the $w$-term collects the coefficient of the mass
moved from $B$ to $\emptyset$. This proves (i): $D_{I,d}$ is linear in
$w$ and strictly negative at $w = 0$ (a Kullback--Leibler divergence
between distinct laws). If $S_{I,d} > 0$ it is strictly increasing and
crosses zero exactly at $w^{*}_{I,d}$, and the displayed bracket is
positive whenever $\aB(\parmw') < \aB(\parmw^{*})$ and $\aN(\parmw') >
\aN(\parmw^{*})$ at the state, the sufficient ordering quoted in the
statement. For (ii), weight each state by its expected clicks under the
plan's myopic occupancy (season-start mass one at $I_0$, depletion with
probability $\ctr{b}\,(\aP + \aB)$, stockout absorbing) and sum: the
season drift is $-\overline{\mathrm{KL}} + w\overline{S}$, with
$\overline{\mathrm{KL}} > 0$ because the two laws differ at some visited
state, and $\overline{S} > 0$ by hypothesis. The hypothesis is a finite
sum of known quantities, so it is checkable at any calibration, and the
pointwise ordering at every visited state is one sufficient condition
for it. Seasons under a frozen plan are i.i.d., and each season's
increment is bounded (finitely many states, all outcome probabilities
positive at visited states by hypothesis), so the strong law of large
numbers gives $n^{-1} \log(L'_n / L^{*}_n) \to -\overline{\mathrm{KL}} +
w\overline{S}$ almost surely. The log-posterior odds differ from
$\log(L'_n / L^{*}_n)$ by the constant log-prior odds, so they diverge
to $-\infty$ below $\bar{w}^{*} = \overline{\mathrm{KL}}/\overline{S}$
and to $+\infty$ above.\qed

\section*{Declarations}
The author declares no known competing financial interests or personal
relationships that could have appeared to influence the work reported in
this paper. This research did not receive any specific grant from funding agencies in
the public, commercial, or not-for-profit sectors.

\paragraph{Data and code availability} All results are simulated from the
parameter values reported in the article and its online supplement. The
code, seeded configurations and result summaries are available at
\url{https://anonymous.4open.science/r/bid-bundle-agent-code-17D7} and
will be deposited in a public repository on acceptance. The online
supplement is included with this preprint as an ancillary file.

\section*{\raggedright Declaration of generative AI and AI-assisted technologies in the manuscript preparation process}
During the preparation of this work the author used Claude (Anthropic) in order to help edit the
text and check the consistency of reported results. After using this tool, the author reviewed and
edited the content as needed and takes full responsibility for the content of the published article.

\setlength{\bibitemsep}{0pt plus 0.3ex}
\printbibliography

\end{document}